\documentclass[a4paper]{article}

\usepackage{INTERSPEECH2021}

\title{Triple M: A Practical Text-to-speech Synthesis System With Multi-guidance Attention And Multi-band Multi-time LPCNet}
\name{Shilun Lin, Fenglong Xie, Li Meng, Xinhui Li, Li Lu}
\address{
  Tencent, Beijing, China}
\email{\{cirolin, fenglongxie, leemeng, hiccupli, adolphlu\}@tencent.com}

\begin{document}

\maketitle
\begin{abstract}
In this work, a robust and efficient text-to-speech (TTS) synthesis system named Triple M is proposed for large-scale online application. The key components of Triple M are: 1) A sequence-to-sequence model adopts a novel multi-guidance attention to transfer complementary advantages from guiding attention mechanisms to the basic attention mechanism without in-domain performance loss and online service modification. Compared with single attention mechanism, multi-guidance attention not only brings better naturalness to long sentence synthesis, but also reduces the word error rate by 26.8\%. 2) A new efficient multi-band multi-time vocoder framework, which reduces the computational complexity from 2.8 to 1.0 GFLOP and speeds up LPCNet by 2.75x on a single CPU. 
\end{abstract}

\noindent\textbf{Index Terms}: Speech synthesis, sequence-to-sequence model, attention, transfer learning, vocoder, LPCNet
\section{Introduction}
In the past few years, speech synthesis has attracted a lot of attention due to advances in deep learning. Sequence-to-sequence neural network \cite{sutskever2014sequence} with attention mechanism is one of the most popular text-to-feature models \cite{wang2017tacotron,shen2018natural}. Attention mechanism is applied to align the input and output sequences. Therefore, the training is no longer fragmented. Content-based attention \cite{bahdanau2015neural} applied in Tacotron \cite{wang2017tacotron} does not exploit the monotonicity and locality between text and acoustic features. Tacotron2 \cite{shen2018natural} attempts to enhance the robustness of attention by introducing location-related information. However, there are still some problems in this hybrid attention mechanism such as lack of monotonic restriction and weak robustness of long sentence synthesis. 
In order to alleviate these problems, alternative attention mechanisms such as forward attention \cite{zhang2018forward}, GMM-based attention \cite{graves2013generating}, stepwise monotonic attention \cite{he2019robust} and dynamic convolution attention \cite{battenberg2020location} are proposed. These attention mechanisms have their own advantages and implementations. It would be nice to transfer those advantages to the basic attention mechanism and keep its original implementation. Knowledge transfer or transfer learning \cite{pan2009survey} between attention mechanisms would be desirable. In this work, we propose a novel multi-guidance attention mechanism which enables the basic location-sensitive attention mechanism to learn different strengths from multiple guidance sources. Forward attention and GMM-based attention are selected as the guiding attention mechanisms after careful analysis. Thanks to the multi-guidance attention mechanism, the convergence speed and the robustness of long sentence synthesis are significantly improved without affecting perceptual quality. More importantly, experiments show that the advantages of guidance attention mechanisms can be transferred to the basic attention through a reasonable combination.

For a large-scale online custom speech synthesis system, it is very important to reduce the cost of  the neural vocoder. The flow based neural vocoders, such as Parallel WaveNet \cite{oord2018parallel} and Clarinet \cite{ping2018clarinet}, reduce the computational overhead of WaveNet \cite{vanwavenet} by learning from the teacher WaveNet and realize real-time synthesis on the GPU. Recently, autoregressive models with simple structures like WaveRNN \cite{kalchbrenner2018efficient} and LPCNet \cite{valin2019lpcnet} have been proposed for real-time synthesis on CPU by introducing sparse GRU. In this work, we propose a multi-band multi-time LPCNet that can simultaneously calculate the excitation of different subbands at adjacent moments in each forward process. The redundant computation is reduced by making full use of the correlation in frequency domain and time domain.

\section{Proposed Method}
\begin{figure*}[!htb]
  \centering
  \includegraphics[scale=0.45]{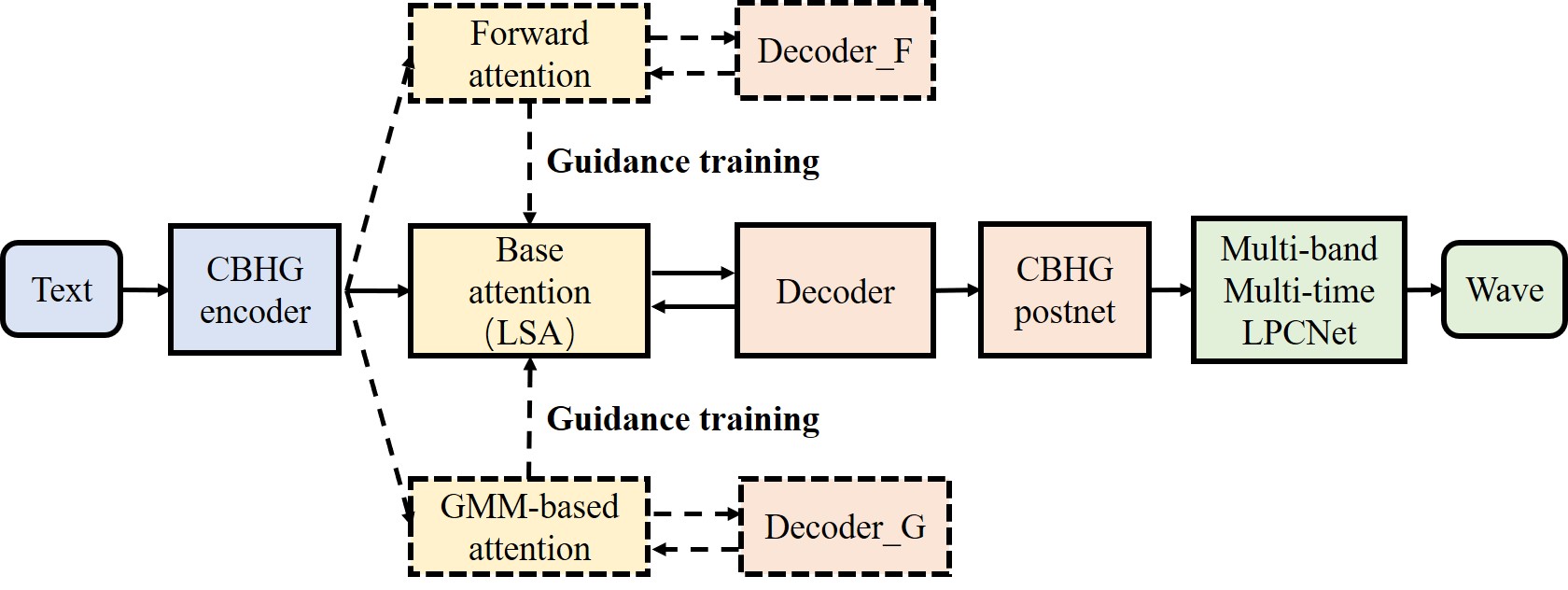}
  \caption{System architecture of our proposed Triple M.}
  \label{fig:triple_m}
\end{figure*}

In this section, we describe main components of triple M as shown in Figure~\ref{fig:triple_m}. First, a text-to-feature module with novel multi-guidance attention will be introduced. The basic location-sensitive attention mechanism learns different strengths, such as fast convergence, stable feature generation and robust long sentence synthesis, from multiple guidance sources during training. In this way, new attention properties which make the system more robust can be integrated without modifying the inference framework of the online system. Next, a novel multi-band multi-time LPCNet which can significantly improve the efficiency of the system will be described. 

\subsection{Multi-guidance attention}
The elaboration of multi-guidance attention will be divided into three parts. Firstly, a basic text-to-feature module which learns from the guidance attention and serves as the final inference module will be described. Then, we will explain why forward and GMM-based attention are chosen as guidance mechanisms. Finally, the details of guidance training will be given.
\subsubsection{Basic setup}
The basic text-to-feature module used in our work is based on Tacotron \cite{wang2017tacotron}. The improved hybrid location-sensitive attention proposed in Tacotron2 \cite{shen2018natural} is applied as the basic attention. CBHG (convolutional bank, highway network and bidirectional gated recurrent unit) encoder is used to transform Chinese Pinyin sequences with tone and prosody information $ \left \{ x_{i}  \right \} _{i=1}^{L}\ $ into hidden text representations $ \left \{ h_{i}  \right \} _{i=1}^{L} $ that are more suitable for attention mechanism (Eq.\ref{1}). An attention RNN uses the last state, context vector and decoding result of the previous time step as input, and outputs the current state $ s_{t} $ for computing the attention score (Eq.\ref{2}). The hybrid location-sensitive attention module takes the current state, hidden representations and location-related information as input to get the attention score $ a_{t} $. And then the context vector $ c_{t} $ at the current moment is calculated (Eq.\ref{3}). Finally, last decoder RNN state, current attention RNN state and the context vector are fed to the decoder RNN. Then the current state of the decoder $ d_{t} $ is obtained and passed to an affine function to obtain the final decoding result $ o_{t} $ (Eq.\ref{4}). 
\begin{equation}\label{1}
\left \{ h_{i}  \right \} _{i=1}^{L} = CBHGEncoder(\left \{ x_{i}  \right \} _{i=1}^{L})
\end{equation}
\begin{equation}\label{2}
s_{t}= AttentionRNN(s_{t-1},c_{t-1},o_{t-1})
\end{equation}
\begin{equation}\label{3}
a_{t}= LSA(s_{t},h_{i},l_{t})  \qquad c_{t}=\sum_{i}^{L} a_{t,i}h_{i}
\end{equation}
\begin{equation}\label{4}
d_{t}= DecoderRNN(d_{t-1},c_{t},s_{t}) \quad o_{t}=Affine(d_{t})
\end{equation}
\subsubsection{Guidance sources selection}
The multi-guidance attention mechanism is introduced into above basic structure and the supervision attention mechanisms are carefully selected. Two complementary attention mechanisms, forward attention \cite{zhang2018forward} and GMM-based attention \cite{graves2013generating} are selected as guidance sources after comparing different attention mechanisms. As a kind of monotonic attention mechanisms, forward attention only considers the alignment paths that satisfy the monotonic condition at each decoding time step to ensure the monotonicity of the final alignment path. This method has been verified to accelerate the convergence and improve the stability of feature generation. However, it is observed that the training process often fails to learn valid alignments. Using a fixed diagonal mask to regulate the attention alignment matrix in training is helpful to solve this problem \cite{lim2020jdi}. The relatively stable basic location-sensitive attention here can play the role of providing diagonal mask. It helps to avoid the collapse of forward attention during training. In turn, the well trained forward attention can give the basic attention an explicit monotonic restriction which is helpful to its guidance learning. For large-scale online applications, we need to improve the robustness of long-form speech synthesis \cite{battenberg2020location} in addition to ensuring the stability of in-domain sentence synthesis. Different from forward attention, GMM-based attention is a purely location-related attention mechanism. It can bring benefits to the basic location-sensitive attention that are different from those of the monotonic attention mechanism. Using GMM-based attention as a guiding mechanism can give the basic location-sensitive attention the ability to synthesize long sentences without modifying the forward computation. This effectively prevents the system from crashing due to long-form input.
\subsubsection{Guidance training}
The guidance attention mechanisms affect the learning of basic attention mechanism through the loss function during training. The training loss function consists of three parts (Eq.\ref{5}). The first part contains $ L_{1} $ distances between all decoder outputs and the real acoustic feature $ r $. $ o, o_{f}, o_{g} $ represent the output of basic decoder, forward attention decoder and GMM-based attention decoder respectively. Then there is the $ L_{1} $ distance between the output of CBHG postnet $ p $ and the real acoustic feature $ r $. The last part $ L\_ga $ (Eq.\ref{6}) includes $ L_{1} $ distances between the basic alignment score $ a $ and all guidance alignment scores. $ a_{f} $ and $ a_{g} $ represent alignment score of forward attention and GMM-based attention. $ \lambda $ is used to control the intensity of guidance learning.
\begin{equation}\label{5}
L=\ell _{1} (o,r)+ \ell _{1} (o_{f},r) +\ell _{1} (o_{g},r) +\ell _{1} (p,r) + L\_ga
\end{equation}
\begin{equation}\label{6}
L\_ga=\lambda(\ell _{1} (a,a_{f})+\ell _{1} (a,a_{g}))
\end{equation}


All guidance-related modules are trained together with the basic text-to-feature module. When the guidance attention mechanisms become reliable, the basic attention mechanism begins to learn from them. Only the basic text-to-feature module is retained in inference.
\subsection{Multi-band multi-time LPCNet}
\subsubsection{LPCNet}
As a relatively lightweight neural vocoder, LPCNet \cite{valin2019lpcnet} models the vocal tract response through a low-cost linear prediction filter and adopts a smaller network to obtain the excitation. However, the computational overhead of LPCNet still has room for improvement. The original LPCNet is composed of a frame rare network (FRN) and a sample rate network (SRN). The former extracts high-level representations from conditional acoustic features. The latter including two GRU layers and a dual FC layer needs to loop T times per frame to obtain T corresponding samples. Therefore, the SRN occupies a major part of the overall calculation.
\subsubsection{Multi-band multi-time processing}
\begin{figure}[ht]
  \centering
  \setlength{\abovecaptionskip}{0.2cm}
  \includegraphics[width=\linewidth]{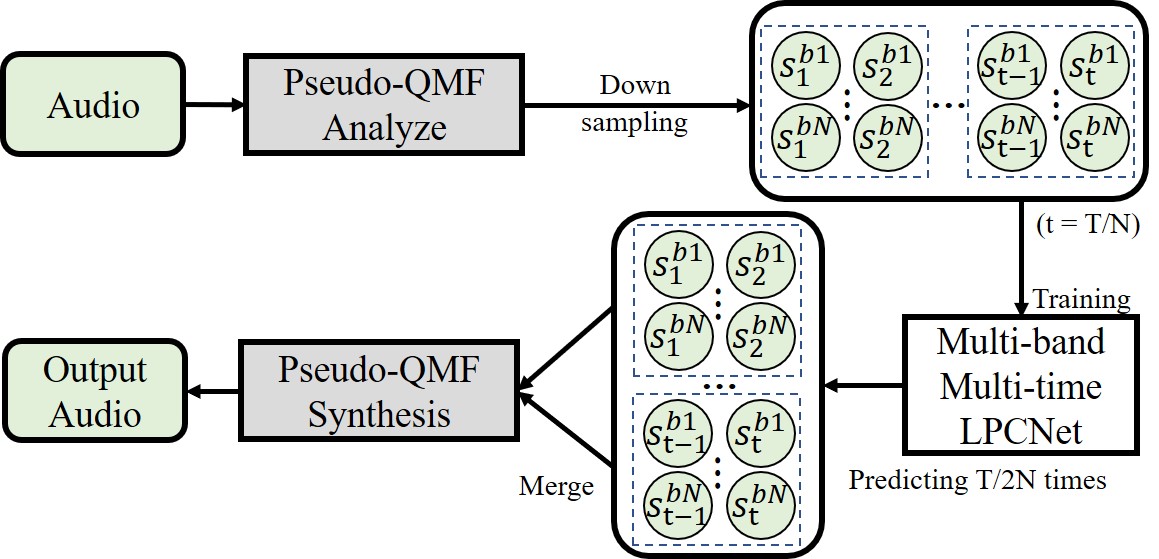}
  \caption{The flowchart of multi-band multi-time processing.}
  \label{fig:mm_processing}
\end{figure}

As analyzed above, the computational complexity of LPCNet is mainly concentrated on SRN. Decreasing the number of SRN forward steps can effectively reduce the computational overhead. Therefore, a multi-band multi-time processing framework is proposed in this work to further explore the acceleration potential of LPCNet by making full use of the correlation in both frequency domain and time domain. The pipeline is shown in Figure~\ref{fig:mm_processing}. 

Multi-band processing is initially used for parallel acceleration of vocoder \cite{okamoto2018improving, okamoto2018investigation}. The speed can be effectively increased by deploying multiple models to calculate different subband signals in parallel, but the computational overhead cannot be truly reduced. Durian proposed in \cite{yu2020durian} introduces multi-band processing into a single WaveRNN model, 
which reduces the total computational complexity from 9.8 to 3.6 GFLOPS. We apply the multi-band strategy to LPCNet for the purpose of further exploiting the latent capacity of SRN. A kind of Cosine-Modulated Filter Bank (CMFB) named Pseudo Quadratue Mirror Filter Bank (Pseudo-QMF) \cite{cruz2002efficient} is applied to multi-band processing. After processing, the original signal is divided into N subbands. Down-sampling each subband N times will not cause the loss of original information. Intuitively, the number of forward steps will be reduced by N times through using SRN to predict N down-sampled subband signals simultaneously. Since Pseudo-QMF is a low-cost filter bank, the cost of reconstructing the original signal from the subband signals is much less than that saved by reducing the number of SRN forward steps. Multi-band strategy improves the efficiency of LPCNet from the frequency domain. The multi-time strategy takes two adjacent sampling points in the subband signal into consideration. SRN predicts adjacent points in N subbands at the same time, which can reduce the number of SRN forward steps by 2N times. Bunched LPCNet \cite{vipperla2020bunched} predicts a bunch of original samples in an autoregressive way. Within a bunch, the first excitation is only conditioned on the output from GRU-B. While for the rest excitation, it depends on all the previous excitation in the same bunch. Different from Bunched LPCNet, our multi-band multi-time LPCNet predicts the adjacent samples of each subband signal at the same time. Due to the small time span, we remove the autoregressive module for simplicity.
\vspace{-0.3cm}

\subsubsection{Audio generation framework}
\begin{figure}[ht]
  \centering
  \setlength{\abovecaptionskip}{0.2cm}
  \includegraphics[width=\linewidth]{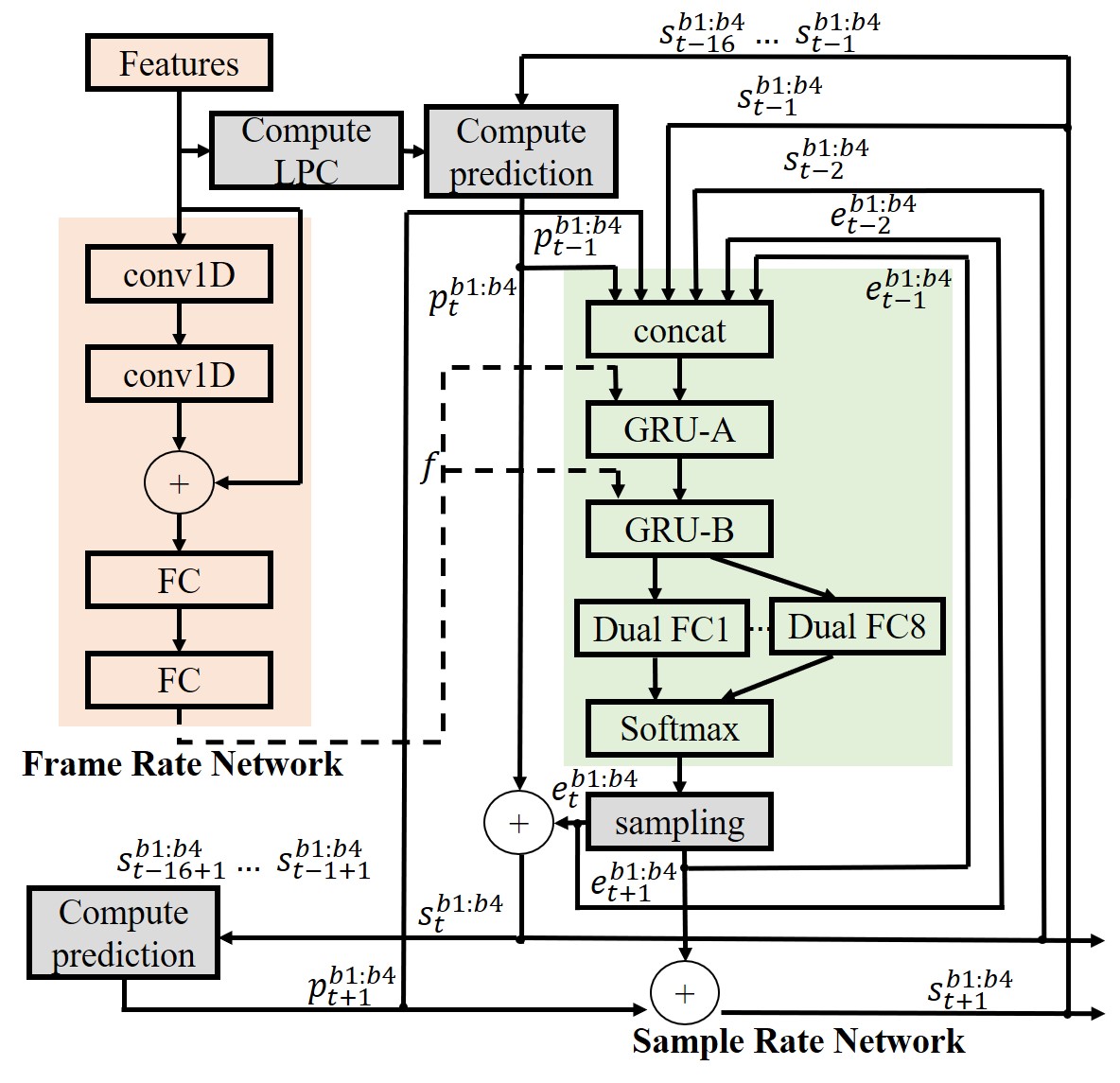}
  \caption{The architecture of multi-band multi-time LPCNet.}
  \label{fig:mm_lpcnet}
\end{figure}

Except for the dual FC layer, the rest of the SRN layers are shared in each forward process as shown in Figure~\ref{fig:mm_lpcnet}. Take four subbands as an example: The excitation ($e_{t-1}^{b1:b4} $ and $e_{t-2}^{b1:b4} $), audio samples ($ s_{t-1}^{b1:b4} $ and $ s_{t-2}^{b1:b4} $) from the last adjacent time and obtained predictions (at the last time $ p_{t-1}^{b1:b4} $ and the current time $ p_{t}^{b1:b4} $) are used as input to the first GRU layer (GRU-A). The output of the second GRU layer (GRU-B) is sent to 8 independent dual FC layers to predict the excitation of subbands at adjacent time ($ e_{t}^{b1:b4} $ and  $ e_{t+1}^{b1:b4} $). After that, the audio samples at the current adjacent time ($ s_{t}^{b1:b4} $ and $ s_{t+1}^{b1:b4} $) can be acquired recursively (Eq.\ref{13},\ref{14},\ref{15}). Finally, the LPC queues of all subbands are updated to prepare for next round. In summary, the input matrix of GRU-A layer will become larger while the table lookup operation makes this overhead negligible. Although 7 more dual FC layers are introduced, the multi-band multi-time strategy reduces the number of required cycles by 8 times which reduces the computational complexity from 2.8 to 1.0 GFLOPS. 
\begin{equation}\label{13}
s_{t}^{b1:b4}=e_{t}^{b1:b4} + p_{t}^{b1:b4} 
\end{equation}
\begin{equation}\label{14}
p_{t+1}^{b1:b4}=LPCPrediction(s_{t-16+1}^{b1:b4}:s_{t-1+1}^{b1:b4})
\end{equation}
\begin{equation}\label{15}
s_{t+1}^{b1:b4}=e_{t+1}^{b1:b4} + p_{t+1}^{b1:b4}
\end{equation}
\vspace{-0.3cm}
\vspace{-0.5cm}

\section{Experiments}
\subsection{Experimental setup}
A Mandarin corpus recorded by a Chinese male speaker was used in our experiments. All recordings were sampled at 16kHz with 16-bit quantization. About 16 hours recordings with an average length of 90 characters were used for training. 100 regular sentences and 50 paragraphs selected from WeChat official account with an average length of 1000 characters were tested. Consistent with original LPCNet, 18 Bark cepstral coefficients and 2 pitch parameters were extracted as conditions of the vocoder. At the same time, they were also used as the prediction targets of the text-to-feature module. 

The text-to-feature module was a sequence-to-sequence model with a CBHG encoder and a CBHG postnet. The main component of decoder was a two-layer 512-dimensional unidirectional LSTM. All attention modules were composed of a unidirectional GRU layer and corresponding attention mechanism. $ \lambda $ was used to control the intensity of guidance learning, which was set to 10 in our experiment. All modules were trained together at the beginning. When the guidance attention 
modules aligned the text and acoustic features stably, the basic attention module began to learn from them. 

In the original LPCNet, SRN consisted of a 90\% sparse 384-dimensional GRU layer, a normal 16-dimensional GRU layer and a 256-dimensional dual FC layer. As to multi-band multi-time LPCNet, dimensions of all GRU layers kept unchanged. Different from \cite{valin2019lpcnet}, condition feature was fed to GRU-B in addition to being input to GRU-A. In order to reach a reasonable compromise between the quality and the computational cost, the hyper-parameters of subband and time span were set to 4 and 2 respectively.

Audio samples related to this work were available  on the accompanying web page\footnote{ https://linshilun.github.io/tripleMsamples/tripleM.html}. 
\subsection{Evaluations}
The experiment found that the basic attention module can align the text and acoustic features stably at the second epoch by learning from forward attention and GMM-based attention. It took much less epochs than 10 epochs required without guidance. This prevented the model falling into bad alignment in early stage of training and provided a stable basis for the subsequent training. The well-learned basic attention provided an approximate diagonal mask which gave a constraint to the forward attention. Thus it was not easy to collapse during training. 
\vspace{-0.3cm}
\begin{table}[th]
  \caption{Long sentence failure rate and word error rate.}
  \vspace{-0.1cm}
  \label{tab:error}
  \centering
  \setlength{\abovecaptionskip}{0.1cm}
  \begin{tabular}{ccc}
    \toprule
    Model & Failure rate & Word error rate \\
    \midrule
    Baseline & \textgreater60\% & NA \\
    Single-guidance attention & \textgreater40\%  & NA \\
    GMM-based attention & 10\%  & 4.1\% \\
    Multi-guidance attention & \textbf{2\%}  & \textbf{3.0\%} \\
    \bottomrule
  \end{tabular}
\end{table}
\vspace{-0.4cm}
\begin{figure}[ht]
  \centering
  \setlength{\abovecaptionskip}{0.2cm}
  \includegraphics[width=\linewidth]{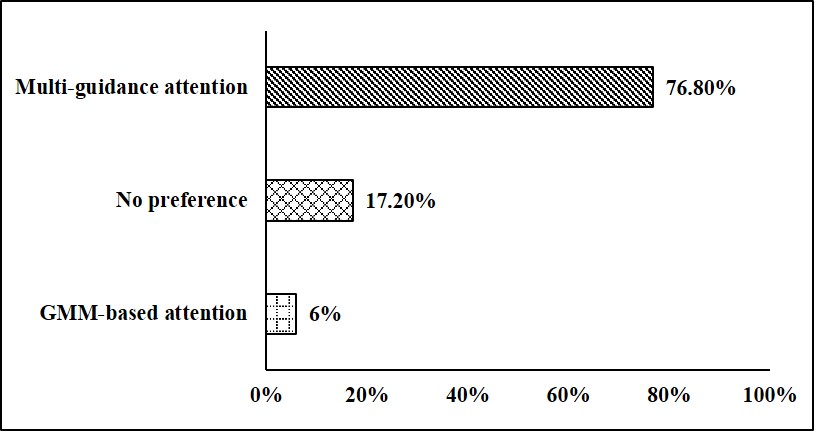}
  \caption{ Multi-guidance attention vs. GMM-based attention, the p-value of t-test between the two systems p\textless0.01.}
  \label{fig:mm_abtext}
\end{figure}
\vspace{-0.2cm}

In the long sentence synthesis experiment, 50 paragraphs selected from WeChat official account covering the fields of politics, sports, entertainment, literature, cooking and so on were tested. The experimental results are recorded in Table~\ref{tab:error}. The failure rates were more than 40\% when forward attention and GMM-based attention were severally used as guidance mechanism. Compared with the baseline model, there was no satisfactory improvement in robustness of long sentences synthesis. The failure was mainly identified by whether the synthesized audio ended early, repeated the same clip or contained meaningless clip which seriously affected the understanding of the content.
Applying two attention mechanisms simultaneously to guide the learning of the basic attention reduced the failure rate to 2\%. These results showed that only using forward attention as a supervisory signal cannot provide long sentence synthesis capabilities. But without the constraint of forward attention, basic attention cannot learn long sentence synthesis ability from GMM-based attention well. Only using GMM-based attention can achieve a 10\% failure rate with 3.3\% in-domain MOS degradation. In the A/B preference test, 10 listeners were asked to listen to the paired sentences synthesized by two different systems in random order and then gave their preference. The result in Figure~\ref{fig:mm_abtext} shows that the audio synthesized by the system with multi-guidance attention was much better. The word error rate (WER) measured by an ASR system described in \cite{wang2020transformer} was 4.1\% when only GMM-based attention was used. Using multi attention mechanisms to guide the learning reduced the WER to 3.0\%. This might be related to the ability of forward attention to generate features stably.

In order to verify the influence of the multi-guidance attention mechanism on perceptual quality, Mean Opinion Score (MOS) tests were implemented. For all MOS tests in this work, three groups of native Chinese speakers (5 in each group) were invited to listen and score 125 audio each time. 100 test utterances synthesized by the corresponding model were mixed with 25 original recordings, and the listener did not know which category each audio belonged to. Scores ranged from 1 to 5, with 5 representing “completely natural speech”. The final MOS was obtained by averaging the scores of the three groups. Table~\ref{tab:MOS1} shows the results of the MOS test. Experiments showed that the basic attention can learn the advantages of the guiding attention while maintaining the in-domain naturalness. In this way, new properties can be assigned to the text-to-speech system without modifying the online service. In addition, multi-guidance attention also provided a new way to improve the performance of attention mechanism through transfer learning.

\vspace{-0.15cm}
\begin{table}[th]
  \caption{MOS with 95\% confidence intervals.}
  \vspace{-0.2cm}
  \label{tab:MOS1}
  \centering
  \setlength{\abovecaptionskip}{0.1cm}
  \begin{tabular}{cc}
    \toprule
     Model & MOS \\
    \midrule
    Baseline        & $ 4.52\pm0.08 $  \\
    GMM-based attention & $ 4.42\pm0.08 $  \\
    Multi-guidance  & $ \textbf{4.57}\pm\textbf{0.05}$    \\
    Ground truth    & $ 4.65\pm0.04 $    \\
    \bottomrule
  \end{tabular}
\end{table}
\vspace{-0.15cm}

According to \cite{valin2019lpcnet}, the complexity of the SRN was around 2.8 GFLOPS. The multi-band multi-time strategy reduced it to around 1.0 GFLOPS (Eq.\ref{16}).
\begin{equation}\label{16}
C=\frac{(3dG_{A}^{2}+3G_{B}(G_{A}+G_{B})+2G_{B}QN_{B}N_{T})2F_{S}}{N_{B}N_{T}}
\end{equation}
\vspace{0.05cm}

Comparison experiment of MOS and real-time factor (RTF) was implemented to demonstrate that the multi-band multi-time strategy can accelerate synthesis without significantly degrading the perceptual quality. The results are presented in Table~\ref{tab:MOS2}. It can be found that the multi-band multi-time strategy sped up LPCNet by 2.75x while the MOS has only dropped by 2.6\%.
\vspace{-0.15cm}
\begin{table}[th]
  \caption{MOS with 95\% confidence intervals and RTF.}
  \vspace{-0.2cm}
  \label{tab:MOS2}
  \centering
  \setlength{\abovecaptionskip}{0.1cm}
  \begin{tabular}{ccc}
    \toprule
    Model & MOS & RTF \\
    \midrule
    LPCNet  & $ 4.57\pm0.05 $ & $ 0.303 $ \\
    Multi-band \& Multi-time & $ \textbf{4.45}\pm\textbf{0.07} $ & $ \textbf{0.110} $  \\
    \bottomrule
  \end{tabular}
\end{table}
\vspace{-0.35cm}
\section{Conclusions}
Robustness and efficiency are essential for a large-scale online system. In this paper, we proposed Triple M, a new practical text-to-speech system. It integrates the advantages of various attention mechanisms by a novel multi-guidance strategy, which enables the system to deal with ultra-long sentence robustly while guaranteeing in-domain performance. Compared with single attention, multi-guidance attention reduces the WER of long sentence synthesis by 26.8\% without MOS degradation. In addition, exploiting the acceleration potential of the original LPCNet through our multi-band multi-time strategy reduces the computational complexity by approximately 64\%. 
\bibliographystyle{IEEEtran}
\bibliography{template}

\end{document}